\documentstyle[aclap]{article}
\title{\vspace{-0.5in}Using Higher-Order Logic Programming for Semantic
Interpretation of Coordinate Constructs}
\author{Seth Kulick \\
University of Pennsylvania \\
Computer and Information Science \\
200 South 33rd Street \\
Philadelphia, PA 19104-6389 USA \\
{\tt skulick@linc.cis.upenn.edu}}

\newcommand{\fa}{\mbox{$>$}}
\newcommand{\ba}{\mbox{$<$}}
\newcommand{\fc}{\mbox{$>B$}}
\newcommand{\bc}{\mbox{$<B$}}
\newcommand{\ft}{\mbox{$>T$}}
\newcommand{\bt}{\mbox{$<T$}}
\newcommand{\lp}{\mbox{$\lambda$Prolog}}
\newcommand{\lt}{\mbox{$\lambda$-term}}
\newcommand{\lts}{\mbox{$\lambda$-terms}}
\begin{document}
\maketitle
\vspace{-0.5in}
\begin{abstract}
Many theories of semantic interpretation use \lt\  manipulation
to compositionally compute the
meaning of a sentence.  These theories are usually implemented in a
language such as Prolog that can simulate \lt\  operations
with first-order unification.  However, for some interesting cases,
such as a Combinatory Categorial Grammar account of coordination
constructs, this can only be done by obscuring the
underlying linguistic theory with the ``tricks'' needed for implementation.
This paper shows how the use of abstract syntax permitted by
higher-order logic programming allows an
elegant implementation of the semantics of Combinatory
Categorial Grammar, including its handling of coordination constructs.
\end{abstract}
\bibliographystyle{acl}
\section{Introduction}
Many theories of semantic interpretation use \lt\  manipulation
to compositionally compute the
meaning of a sentence.  These theories are usually implemented in a
language such as Prolog that can simulate \lt\  operations
with first-order unification.  However, there are cases in which this
can only be done by obscuring the
underlying linguistic theory with the ``tricks'' needed for implementation.
For example,
Combinatory Categorial Grammar (CCG) \cite{Steedman90} is a theory of
syntax and semantic interpretation that has the attractive characteristic
of handling many coordination constructs that other theories cannot.
While many aspects of CCG semantics can
 be reasonably simulated in first-order unification, the
simulation breaks down on some of the most interesting cases that CCG
can theoretically handle.  The problem in general, and for CCG in particular,
is that the implementation language does not have sufficient
expressive power to allow a more direct encoding.  The solution given in this
paper is to show how
advances in logic programming allow the implementation
of semantic theories in a very direct and natural way, using CCG
as a case study.


We begin by briefly illustrating
why first-order unification is inadequate for some
coordination constructs, and then review two proposed solutions. The sentence
in (1a) usually has the logical form (LF) in (1b).
\begin{enumerate}
\item[(1a)] \verb+John and Bill run.+
\item[(1b)] \verb+(and (run John) (run Bill))+
\end{enumerate}
CCG is one of several theories in which (1b) gets
derived by raising {\it John} to be the LF
$\lambda P.(P~{\tt john})$, where
$P$ is a predicate that takes a NP as an argument to return a sentence.
Likewise, {\it Bill} gets the LF $\lambda P.(P~{\tt bill})$,
and coordination
results in the following LF for {\it John and Bill}:
\begin{enumerate}
\item[(2)] $\lambda P.({\tt and}~(P~{\tt john})~(P~{\tt bill}))$
\end{enumerate}
When (2) is applied to the predicate, (1b) will
result after $\beta$-reduction.  However, under first-order unification,  this
needs to simulated by having the variable $x$ in $\lambda x.run(x)$
unify both with Bill and John, and this is not possible.  See
\cite{Jowsey} and \cite{Moore:ACL89} for a thorough discussion.

\cite{Moore:ACL89} suggests that the way to overcome this problem is to
use explicit \lts\ and encode
$\beta$-reduction to perform the needed reduction.  For example, the
logical form in (3) would be produced, where
${\tt X}\backslash{\tt run(X)}$ is the
representation of $\lambda x.{\tt run}~(x)$.
\begin{enumerate}
\item [(3)] \verb+and(apply(X\run(X),john),+
\item []    \verb+    apply(X\run(X),bill))+
\end{enumerate}
This would then be reduced by the clauses for {\tt apply} to result in (1b).
For this small example, writing such an {\tt apply} predicate is
not difficult.
However, as the semantic terms become more complex, it is no
trivial matter to write $\beta$-reduction that will correctly handle
variable capture.  Also, if at some point it was desired to determine if
the semantic forms of two different sentences were the same, a predicate
would be needed to compare two lambda forms for $\alpha$-equivalence,  which
again is not a simple task.  Essentially, the logic variable {\tt X} is
meant to be interpreted as a bound variable, which requires an additional layer
of programming.

\cite{Park:ACL92} proposes a solution within first-order unification that can
handle not only sentence (1a), but also more complex examples with
determiners.  The method used is to
introduce spurious bindings that subsequently
get removed.  For example, the semantics of (4a) would be (4b), which would
then get simplified to (4c).
\begin{enumerate}
\item [(4a)] A farmer and every senator talk
\item [(4b)]
\verb+exists(X1,farmer(X1)+ \\
\verb+&(exists(X2,(X2=X1)&talk(X2))))+ \\
\verb+&forall(X3,senator(X3)+ \\
\verb+=>(exists(X2,(X2=X3)&talk(X2))))+
\item [(4c)]
\verb+exists(X1,farmer(X1)&talk(X1))+ \\
\verb+&forall(X3,senator(X3)=>talk(X3))+
\end{enumerate}

While this pushes first-order unification beyond what it had been previously
shown capable of, there are two disadvantages to this technique:
(1) For every possible category that can be conjoined, a separate
lexical entry for {\it and} is required, and (2)
As the conjoinable categories become more complex, the {\it and} entries
become correspondingly more complex and greatly
obscure the theoretical background of the grammar formalism.

The fundamental problem in both cases is that the concept of free and bound
occurrences of variables is not supported by Prolog, but instead needs to
be implemented by additional programming.  While theoretically possible,
it becomes quite problematic to actually implement.  The solution given
in this paper is to use a higher-order logic programming language, \lp,
that already implements these concepts, called ``abstract syntax'' in
\cite{Miller:Abs91} and ``higher-order abstract syntax'' in \cite{Pfen+El}.
This allows a natural and elegant implementation of the grammatical theory,
with only one  lexical entry for {\it and}.  This paper is meant to
be viewed as furthering the exploration of the utility of higher-order
logic programming for computational linguistics - see, for example,
\cite{MN}, \cite{Pareschi}, and \cite{Pereira}.

\section{CCG}
CCG is a grammatical formalism in which there is a one-to-one correspondence
between the rules of composition\footnote{In the general sense, not
specifically the CCG rule for function composition.}
at the level of syntax and logical form. Each word is
(perhaps ambiguously) assigned
a category and LF, and when the syntactical operations assign a new
category to a constituent, the corresponding semantic operations
produce a new LF for that constituent as well.  The CCG rules
\begin{figure}
\begin{enumerate}
\item [] Function Application (\fa): \\
${\tt X/Y:F~~Y:y~=>X:Fy}$
\item [] Function Application (\ba): \\
${\tt Y:y~~X\backslash Y:F=>X:Fy}$
\item [] Function Composition (\fc): \\
${\tt X/Y:F~~Y/Z:G=>X/Z:\lambda x.F(Gx)}$
\item [] Function Composition (\bc): \\
${\tt Y\backslash Z:G~~X\backslash Y:F=>X\backslash Z:\lambda x.F(Gx)}$
\item [] Type Raising (\ft): \\
${\tt np:x=>s/(s\backslash np):\lambda F.Fx}$
\item [] Type Raising (\bt): \\
${\tt np:x~=>s\backslash (s/np):\lambda F.Fx}$
\end{enumerate}
\caption{CCG rules}
\end{figure}
shown in Figure 1 are implemented in the system described in this
paper.\footnote{The type-raising rules shown are actually a simplification
of what has been implemented.  In order to handle determiners, a system
similar to NP-complement categories as discussed in \cite{Dowty} is used.
Although a worthwhile further demonstration of the use of abstract syntax, it
has been left out of this paper for space reasons.} \footnote{The
$\backslash$ for a backward-looking category should not be confused with the
$\backslash$ for $\lambda$-abstraction.} Each of the
three operations have both a forward and backward variant.

As an illustration of how the semantic rules can be simulated in first-order
unification, consider the derivation of the constituent {\it harry found},
where {\it harry} has the category {\tt np} with LF {\tt harry'} and
{\it found} is a transitive verb of category
${\tt (s\backslash np)/np}$ with LF
\begin{enumerate}
\item [(5)] $\lambda object.\lambda subject.({\tt found'}~subject~object)$
\end{enumerate}
\begin{figure}
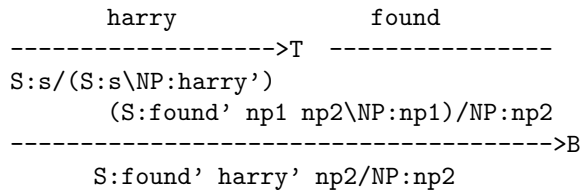

\label{harryfound}
\begin{verbatim}
       harry              found
------------------->T  ----------------
S:s/(S:s\NP:harry')
       (S:found' np1 np2\NP:np1)/NP:np2
--------------------------------------->B
      S:found' harry' np2/NP:np2
\end{verbatim}
\caption{CCG derivation of {\it harry found} simulated by first-order
unification}
\end{figure}
In the CCG formalism, the derivation is as follows: {\it harry}
gets raised with the \ft\ rule, and then forward composed by the \fc\ rule
with {\it found}, and the result is a category of type
{\tt s/np} with
LF $\lambda x.({\tt found'~harry'}~x)$.  In section 3 it will be
seen how the use of abstract syntax allows this to be expressed directly.
In first-order unification, it is simulated as shown in
Figure 2.\footnote{example adapted from
\cite[p. 220]{Steedman90}.}

The final CCG rule to be considered is the coordination rule that specifies
that only like categories can coordinate:
\begin{enumerate}
\item [(6)] \verb+X conj X => X+
\end{enumerate}
This is actually a schema for a family of rules, collectively called
``generalized coordination'', since the semantic rule is different
for each case.\footnote{It is not established if this schema should actually
produce an unbounded family of rules.  See \cite{Weir} and
\cite{Weir and Joshi} for a
discussion of the implications for automata-theoretic
power of generalized coordination
and composition, and \cite{Gazdar} for linguistic arguments that languages
like Dutch may require this power, and \cite{Steedman90} for some further
discussion of the issue.  In this paper we use the generalized rule to
illustrate the elegance of the representation, but it is an easy
change to implement a bounded coordination rule.}
For example, if {\tt X} is a unary function,
then the semantic rule is (7a), and
if the functions have two arguments, then the rule is
(7b).\footnote{The $\Phi$ notation is used because of the combinatory logic
background of CCG.  See \cite{Steedman90} for details.}
\begin{enumerate}
\item [(7a)] $\Phi FGH=\lambda x.F(Gx)(Hx)$
\item [(7b)] $\Phi^2 FGH=\lambda x.\lambda y.F(Gxy)(Hxy) $
\end{enumerate}
For example, when processing (1a), rule (7a) would be used with:
\begin{itemize}
\item $F=\lambda x.\lambda y.({\tt and'}~x~y)$
\item $G=\lambda P.(P~{\tt john'})$
\item $H=\lambda P.(P~{\tt bill'})$
\end{itemize}
with the result
\begin{displaymath}
\phi FGH= \lambda x.({\tt and'}~(x~{\tt john'})~(x~{\tt bill'}))
\end{displaymath}
which is $\alpha$-equivalent to (2).

\section{$\lambda$PROLOG and Abstract Syntax}
\lp\ is a logic programming language based on
{\it higher-order hereditary Harrop formulae} \cite{MNPS}.  It differs from
Prolog in that first-order terms and unification are replaced with
simply-typed $\lambda$-terms and higher-order
 unification\footnote{defined as the unification of simply typed \lts,
modulo $\beta\eta$ conversion.}, respectively.  It also permits universal
quantification and implication in
the goals of clauses. The crucial aspect for this paper is that together
these features permits the usage of abstract syntax to express the logical
forms terms computed by CCG.  The built-in $\lambda$-term manipulation is
used as a ``meta-language'' in which the ``object-language'' of CCG
logical forms is expressed, and variables in the object-language are mapped
to variables in the meta-language.

\begin{figure}
\label{lp1}
\begin{verbatim}
kind    tm      type.
type    abs     (tm -> tm) -> tm.
type    app     tm -> tm -> tm.

type    forall  (tm -> tm) -> tm.
type    exists  (tm -> tm) -> tm.
type    >>      tm -> tm -> tm.
type    &&      tm -> tm -> tm.
\end{verbatim}
\caption{Declarations for \lp\ representation of CCG logical forms}
\end{figure}

The \lp\ code fragment shown in Figure 3 declares how the CCG
logical forms are represented.  Each CCG LF is represented as
an untyped $\lambda$-term, namely type {\tt tm}.  {\tt abs} represents
object-level abstraction $\lambda x.M$ by the meta-level expression
{\tt (abs N)}, where N is a meta-level function of type
${\tt tm \rightarrow tm}$.  A meta-level $\lambda$-abstraction $\lambda y.P$
is written ${\tt y\backslash P}$.\footnote{\label{fnhere}This is the same
syntax for
$\lambda$-abstraction as in (3).  \cite{Moore:ACL89} in fact borrows the
notation for $\lambda$-abstraction from \lp. The difference, of course,
is that here the abstraction is a meta-level, built-in construct, while
in (3) the interpretation is dependent on an extra layer of programming.
Bound variables in \lp\ can be either upper or lower case, since they
are not logic variables, and will be written in lower case in this
paper.}

  Thus, if {\tt walked'} has type ${\tt tm \rightarrow tm}$,
then ${\tt y\backslash (walked'~y)}$ is a \lp\ (meta-level) function with type
${\tt tm \rightarrow tm}$, and ${\tt (abs~y\backslash (walked'~y))}$
is the object-level
representation, with type {\tt tm}.  The LF for
{\it found} shown in (5) would be represented as
${\tt (abs~obj\backslash (abs~sub\backslash (found'~sub~obj)))}$.  {\tt app}
 encodes application, and so in the derivation of
{\it harry found}, the
type-raised {\it harry} has the \lp\ value
${\tt (abs~p\backslash (app~p~harry'))}$.\footnote{It is possible to represent
the logical forms at the object-level without using {\tt abs} and {\tt app},
so that {\it harry}  could
be simply ${\tt p\backslash (p~harry')}$. The original
implementation of this
system was in fact done in this manner.  Space prohibits a full explanation,
but essentially the fact that
\lp\ is a typed language leads to a good deal of formal clutter if
this method is used.}

The second part of Figure 3 shows declares how quantifiers are
represented, which are required since the sentences to be processed
may have determiners.  {\tt forall} and {\tt exists} are encoded
similarly to abstraction, in that they take a functional argument and
so object-level binding of variables by quantifiers is handled by
meta-level $\lambda$-abstraction.  \verb+>>+ and \verb+&&+ are
simple constructors for implication and conjunction, to be used with
{\tt forall} and {\tt exists} respectively, in the typical
manner \cite{P+S}.  For example, the sentence
{\it every man found a bone} has as a possible LF (8a), with
the \lp\ representation (8b)\footnote{\label{ccgfn}The LF for the
determiner has the form of a Montagovian generalized quantifier,
giving rise to one fully scoped logical form for the sentence.
It should
be stressed that this particular kind of LF is assumed here
purely for the sake
of illustration, to make the point that composition at the level of
derivation and LF are one-to-one.  Section 4 contains an example
for which such a derivation fails to yield all available quantifier scopings.
We do not address here the further question of how the remaining scoped
readings are derived.  Alternatives that appear compatible with the
present approach are quantifier movement \cite{HS}, type-raising at LF
\cite{PR}, or the use of disambiguated quantifers in the derivation itself
\cite{Park:ACL95}.}:
\begin{enumerate}
\item [(8a)]
$\exists x.(({\tt bone'}~x)~\wedge~\forall y.(({\tt man'}~y) \rightarrow
({\tt found'}~y~x)))$
\item [(8b)]
\begin{verbatim}
(exists x\
  ((bone' x) &&
   (forall x1\
     ((man' x1) >> (found' x1 x)))))
\end{verbatim}
\end{enumerate}

\begin{figure}
\label{lptwo}
\begin{verbatim}
type	apply   tm -> tm -> tm -> o.
type	compose tm -> tm -> tm -> o.
type	raise   tm -> tm -> o.

apply   (abs R) S (R S).
compose	(abs F) (abs G) (abs x\(F (G x))).
raise   Tm (abs P\(app P Tm)).
\end{verbatim}
\caption{\lp\ implementation of CCG logical form operations}
\end{figure}

Figure 4 illustrates how directly the CCG operations can be
encoded\footnote{There are other clauses, not shown here, that determine
the direction
of the CCG rule.  For either direction, however, the semantics are the
same and both directional rules call these clauses for the semantic
computation.}. {\tt o} is the type of a meta-level proposition, and
so the intended usage of {\tt apply} is to take three arguments of type
{\tt tm}, where the first should be an object-level $\lambda$-abstraction,
and set the third equal to the application of the first to the second.
Thus, for the query
\begin{verbatim}
?- apply (abs sub\(walked' sub)) harry' M.
\end{verbatim}
{\tt R} unifies with the ${\tt tm \rightarrow tm}$
function ${\tt sub\backslash (walked'~sub)}$, S with {\tt harry'} and
M with {\tt (R S)}, the meta-level application of R to S, which by the
built-in $\beta$-reduction is {\tt (walked' harry')}.  In other words,
 object-level function application is handled simply by the
meta-level function application.

Function composition is similar. Consider again the derivation
of {\it harry} {\it found} by type-raising and forward composition.
{\it harry}
would get type-raised by the {\tt raise} clause to produce
${\tt (abs~p\backslash (app~p~harry'))}$,
and then composed with {\it found}, with
the result shown in the following query:
\begin{verbatim}
?- compose (abs p\(app p harry'))
           (abs obj\
            (abs sub\
             (found' sub obj)))
            M.
M = (abs x\
     (app
      (abs sub\(found' sub x))
      harry')).
\end{verbatim}
At this point a further $\beta$-reduction is needed.
Note however this is not at all the same problem of writing a
$\beta$-reducer in Prolog.  Instead it is a simple matter of using
the meta-level $\beta$-reduction to eliminate $\beta$-redexes to
produce the final result ${\tt (abs~x\backslash (found'~harry~x))}$.
We won't show the complete declaration of the $\beta$-reducer, but
the key clause is simply:
\begin{verbatim}
red	(app (abs M) N) (M N).
\end{verbatim}

Thus, using the abstract syntax capabilities of \lp, we can have a
direct implementation of the underlying linguistic formalism, in stark
contrast to the first-order simulation shown in Figure 2.
\section{Implementation of Coordination}
A primary goal of abstract-syntax is to support recursion through
abstractions with bound variables.  This leads to
the interpretation
of a bound variable as a ``scoped constant'' - it acts like a constant
that is not visible from the top of the term, but which becomes visible
during the descent through the abstraction.  See \cite{Miller:Abs91}
for a discussion of how this may be used for evaluation of functional
programs by ``pushing'' the evaluation through abstractions to reduce
redexes that are not at the top-level.  This technique is also used in
the $\beta$-reducer briefly mentioned at the end of the previous section,
and a similar technique will be used here to implement coordination by
recursively descending through the two arguments to be coordinated.
\begin{figure}
\label{ccgtypes}
\begin{verbatim}
kind    cat             type.
type    fs              cat -> cat -> cat.
type    bs              cat -> cat -> cat.

type    np              cat.
type    s               cat.
type    conj            cat.
type    noun            cat.

type    atomic-type     cat -> o.

atomic-type np.
atomic-type s.
atomic-type conj.
atomic-type noun.

\end{verbatim}
\caption{Implementation of the CCG category system}
\end{figure}

\begin{figure}
\label{coord}
\begin{verbatim}
type    coord
          cat -> tm -> tm -> tm -> o.

coord (fs A B) (abs R) (abs  S) (abs T) :-
        pi x\ (coord B (R x) (S x) (T x)).

coord (bs A B) (abs R) (abs  S) (abs T) :-
        pi x\ (coord B (R x) (S x) (T x)).

coord B R S (and' R S) :- atomic-type B.
\end{verbatim}
\caption{Implementation of coordination}
\end{figure}

Before describing the implementation of coordination, it is
first necessary to mention how CCG categories are
represented in the \lp\ code.  As shown in Figure
5, {\tt cat} is declared to be a primitive
type, and {\tt np}, {\tt s}, {\tt conj}, {\tt noun}
are the categories used in this implementation.
{\tt fs} and {\tt bs} are declared to be constructors for
forward and backward slash.  For example, the CCG
category for a transitive verb ${\tt (s\backslash np)/np}$ would be
represented as {\tt (fs np (bs np s))}.   Also, the
predicate {\tt atomic-type} is declared to be true
for the four atomic categories.  This will be used in the
implementation of coordination as a test for termination of
the recursion.

The implementation of coordination crucially uses the
capability of \lp\ for universal quantification in
the goal of a clause.  {\tt pi} is the
meta-level operator for $\forall$, and $\forall x.M$ is written as
${\tt pi~x\backslash M}$.  The operational semantics for \lp\ state that
${\tt pi~x\backslash G}$ is provable if and only if $[c/x]G$ is
provable, where $c$ is a new variable of the same type as $x$ that
does not otherwise occur in the current signature.  In other words,
$c$ is a scoped constant and the current signature gets
expanded with $c$ for the proof of $[c/x]G$.  Since $c$ is
meant to be treated as a generic placeholder for any arbitrary
$x$ of the proper type, $c$ must not appear in
any terms instantiated for logic variables during the proof of
$[c/x]G$.  The significance of this restriction will be illustrated shortly.

The code for coordination is shown in Figure 6.
The four arguments to {\tt coord} are a category and
three terms that are the object-level
LF representations of constituents of that category.
The last argument will result from the coordination of the
second and third arguments.  Consider again the earlier problematic
example (1a) of coordination.  Recall that after {\it john} is type-raised,
its LF will be ${\tt (abs~p\backslash (app~p~john'))}$ and similarly
for {\it bill}.  They will both have the category {\tt (fs (bs np s) s)}.
Thus, to obtain the LF for {\it John and Bill}, the following
query would be made:
\begin{verbatim}
?- coord (fs (bs np s) s)
         (abs p\(app p john'))
         (abs p\(app p bill'))
         M.
\end{verbatim}
This will match with the first clause for \verb+coord+, with
\begin{itemize}
\item {\tt A} instantiated to {\tt (bs np s)}
\item {\tt B} to {\tt s}
\item {\tt R} to ${\tt (p\backslash (app~p~john'))}$
\item {\tt S} to ${\tt (p\backslash (app~p~bill'))}$
\item and {\it T} a logic variable waiting instantiation.
\end{itemize}
  Then, after the meta-level
$\beta$-reduction using the new scoped constant {\tt c},
the following goal is called:
\begin{verbatim}
?- coord s (app c john') (app c bill') N.
\end{verbatim}
where {\tt N = (T c)}.  Since {\tt s} is an atomic type, the third
{\tt coord} clause matches with
\begin{itemize}
\item {\tt B} instantiated to {\tt s}
\item {\tt R} to {\tt (app c john')}
\item {\tt S} to {\tt (app c bill')}
\item {\tt N} to {\tt (and' (app c john') (app c bill'))}
\end{itemize}
 Since {\tt N = (T c)}, higher-order unification is used by \lp\ to instantiate
{\tt T} by extracting {\tt c} from {\tt N} with the result
\begin{displaymath}
{\tt T = x\backslash (and'~(app~x~john')~(app~x~bill'))}
\end{displaymath}
and so {\tt M} from the original query is
\begin{displaymath}
{\tt (abs~x\backslash(and'~(app~x~john')~(app~x~bill')))}
\end{displaymath}
Note that since
{\tt c} is a scoped constant arising from the proof of an universal
quantification, the instantiation
\begin{displaymath}
{\tt T = x\backslash (and'~(app~c~john')~(app~x~bill'))}
\end{displaymath}
is prohibited, along with
the other extractions that do not remove {\tt c} from the body of
the abstraction.

This use of universal quantification to extract out $c$ from a term containing
$c$ in this case gives the same result as a direct implementation of
the rule for cooordination of unary functions (7a) would.  However,
this same process of recursive
descent via scoped constants will work for any member of the
{\tt conj} rule family.  For example, the following query
\begin{verbatim}
?- coord
    (fs np (bs np s))
    (abs obj\(abs sub\(like' sub obj)))
    (abs obj\(abs sub\(hate' sub obj)))
    M.
M = (abs x\
     (abs x1\
      (and' (like' x1 x)
            (hate' x1 x)))).
\end{verbatim}
corresponds to rule (7b).  Note also that the use of the
same bound variable names {\tt obj} and {\tt sub} causes no difficulty
since the use of scoped-constants, meta-level $\beta$-reduction, and
higher-order unification is used to access and manipulate the
inner terms.   Also, whereas \cite{Park:ACL92} requires careful
consideration of handling of determiners with coordination, here
such sentences are handled just like any others.  For example,
the sentence {\it Mary gave every dog a bone and some policeman
a flower} results in the LF
\footnote{This is a case in which the particular LF assumed here
fails to yield another available scoping.  See footnote~\ref{ccgfn}.}:
\begin{verbatim}
(and'
 (exists x\((bone' x) &&
     (forall x1\((dog' x1)
          >> (gave' mary' x x1)))))
 (exists x\((flower' x) &&
     (exists x1\((policeman' x1)
          && (gave' mary' x x1))))))
\end{verbatim}
Thus, ``generalized coordination'', instead of being a family of separate
rules, can be expressed as a single rule on recursive descent through
logical forms.  \cite{Steedman90} also discusses
``generalized  composition'',
and it may well be that a similar implementation is possible for that
family of rules as well.

\section{Conclusion}
We have shown how higher-order logic programming
can be used to elegantly implement the semantic theory of CCG, including
the previously difficult case of its handling of coordination constructs.
The techniques used here should allow similar advantages for a variety
of such theories.

An argument can be made that the approach taken here relies on
a formalism that entails implementation issues that are more difficult
than for the other solutions and inherently not as efficient.
However, the implementation issues, although more complex, are also
well-understood  and it can be expected
that future work will bring further improvements.  For example, it is
a straightforward matter to transform the \lp\ code into a logic
called $L_\lambda$ \cite{Miller:ll} which requires only a restricted form of
unification that is decidable in linear time and space.  Also, the
declarative nature of \lp\ programs opens up the possibility for applications
of program transformations such as partial evaluation.

\section{Acknowledgments}

This work is supported by ARO grant DAAL03-89-0031,
DARPA grant N00014-90-J-1863, and ARO grant DAAH04-94-G-0426.  I would like
to thank Aravind Joshi, Dale Miller, Jong Park, and Mark Steedman for
valuable discussions and comments on earlier drafts.

\end{document}